\documentclass[lettersize,journal]{IEEEtran}
\usepackage{amsmath,amsfonts}
\usepackage{algorithmic}
\usepackage{algorithm}
\usepackage{array}
\usepackage{textcomp}
\usepackage{subcaption}
\usepackage{hyperref}
\usepackage{stfloats}
\usepackage{booktabs}
\usepackage{url}
\usepackage{verbatim}
\usepackage{multirow}
\usepackage{graphicx}
\usepackage{cite}
\usepackage{arydshln}
\usepackage{xcolor}
\usepackage{xpatch}
\usepackage{overpic}
\usepackage{color}
\usepackage{hyperref}
\usepackage{algorithm}
\usepackage{algorithmic}
\newlength\savewidth
\newcommand{\tablestyle}[2]{\setlength{\tabcolsep}{#1}\renewcommand{\arraystretch}{#2}\centering\footnotesize}
\renewcommand{\paragraph}[1]{\vspace{1.25mm}\noindent\textbf{#1}}

\hypersetup{
    colorlinks=true,
    linkcolor=blue,
    filecolor=blue,      
    urlcolor=blue,
    citecolor=blue,
}
\begin{document}

\title{Semantic-Guided Global-Local Collaborative Networks for Lightweight Image Super-Resolution}

\author{Wanshu Fan, Yue Wang, Cong Wang, Yunzhe Zhang, Wei Wang and Dongsheng Zhou\

\thanks{This work was supported in part by the National Key Research and Development Program of China (Grant No. 2021ZD0112400), Centre for Advances in Reliability and Safety (CAiRS) admitted under AiR@InnoHK Research Cluster, National Natural Science Foundation of China (Grant No. U1908214), the Program for Innovative Research Team in University of Liaoning Province (Grant No. LT2020015), the Support Plan for Key Field Innovation Team of Dalian (2021RT06), the Support Plan for Leading Innovation Team of Dalian University (XLJ202010), 
Interdisciplinary project of Dalian University (Grant No. DLUXK-2023-QN-015). (Corresponding author: Dongsheng Zhou)}
\thanks{Wanshu Fan, Yue Wang Yunzhe Zhang and Dongsheng Zhou are with the National and Local Joint Engineering Laboratory of Computer Aided Design, School of Software Engineering, Dalian University, Dalian, China (E-mail: fanwanshu@dlu.edu.cn, wangyue@s.dlu.edu.cn, 
zhangyunzhe@s.dlu.edu.cn,
zhouds@dlu.edu.cn).}

\thanks{Cong Wang is with the Centre for Advances in Reliability and Safety (CAiRS), Hong Kong, China (E-mail: supercong94@gmail.com).}
\thanks{Wei Wang is with School of Cyber Science and Technology, Sun Yat-Sen University, Shen Zhen, China  (E-mail: wangwei29@mail.sysu.edu.cn).}
}


\date{\footnotesize\textsuperscript{\textbf{1}}National and Local Joint Engineering Laboratory of Computer Aided Design,\\
School of Software Engineering, Dalian University, Dalian 116622, LiaoNing, China\\ }
\markboth{Journal of \LaTeX\ Class Files,~Vol.~14, No.~8, August~2021}%
{Shell \MakeLowercase{\textit{et al.}}: A Sample Article Using IEEEtran.cls for IEEE Journals}


\maketitle

\begin{abstract}
Single-Image Super-Resolution (SISR) plays a pivotal role in enhancing the accuracy and reliability of measurement systems, which are integral to various vision-based instrumentation and measurement applications.
These systems often require clear and detailed images for precise object detection and recognition.
However, images captured by visual measurement tools frequently suffer from degradation, including blurring and loss of detail, which can impede measurement accuracy.
%
As a potential remedy, we in this paper propose a Semantic-Guided Global-Local Collaborative Network (SGGLC-Net) for lightweight SISR. Our SGGLC-Net leverages semantic priors extracted from a pre-trained model to guide the super-resolution process, enhancing image detail quality effectively.
Specifically, we propose a Semantic Guidance Module that seamlessly integrates the semantic priors into the super-resolution network, enabling the network to more adeptly capture and utilize semantic priors, thereby enhancing image details.
To further explore both local and non-local interactions for improved detail rendition, we propose a Global-Local Collaborative Module, which features three Global and Local Detail Enhancement Modules, as well as a Hybrid Attention Mechanism to work together to efficiently learn more useful features.
%
Our extensive experiments show that SGGLC-Net achieves competitive PSNR and SSIM values across multiple benchmark datasets, demonstrating higher performance with the multi-adds reduction of 12.81G compared to state-of-the-art lightweight super-resolution approaches.
These improvements underscore the potential of our approach to enhance the precision and effectiveness of visual measurement systems.
Codes are at \href{https://github.com/fanamber831/SGGLC-Net}{https://github.com/fanamber831/SGGLC-Net}.

\end{abstract}

\begin{IEEEkeywords}
Lightweight image super-resolution, vision-based measurement, semantic guidance module, global-local collaborative module.
\end{IEEEkeywords}

\section{Introduction}
\IEEEPARstart{S}{ingle}-image Super-Resolution (SISR) stands as a critical and complex challenge within the field of computer vision, with significant implications for instrumentation and measurement applications.
Its primary objective is to reconstruct a high-resolution (HR) image from a corresponding low-resolution (LR) version, which is essential for precise visual measurement tasks. 
Beyond its inherent complexities, SISR has a myriad of applications spanning diverse vision-based measurement activities critical to fields such as medical imaging, remote sensing, and surveillance to more advanced vision-based measurement tasks like object detection and Image segmentation.

The ability of SISR to enhance image quality is particularly important for measurement systems operating under challenging conditions, such as low light or atmospheric disturbances. By increasing image resolution, SISR helps ensure that these systems maintain their robustness and reliability. Additionally, SISR can reduce costs by enabling the use of lower-cost imaging equipment while still achieving high-resolution results through post-processing, thus making measurement systems more efficient and extending the lifespan of existing equipment. Despite its benefits, the SISR problem is intrinsically ill-posed, as multiple HR images can correspond to the same LR input, highlighting the complexity and significance of SISR in enhancing measurement and analysis across various fields.

\begin{figure}[!t] 
\centering
\includegraphics[width=0.95\linewidth]{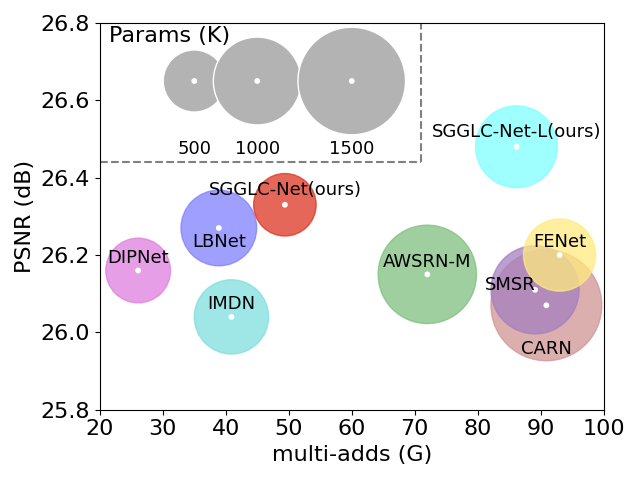}
\caption{Model multi-adds comparison on Urban100~\cite{huang2015single} ($\times$4), where the output image size is 1280$\times$720. 
Our SGGLC-Net family achieves a better trade-off between model complexity and super-resolution performance.
}
\label{fig:p7}
\end{figure}

In addressing this issue, researchers have extensively explored deep neural network-based super-resolution methods \cite{0.5,0.6,0.7,0.8}, achieving commendable results in image reconstruction. Structure-Aware Deep Network~\cite{0.5}, the structure perception module enhances the model's ability to recover geometric structures in images, improving super-resolution visual quality. Generative adversarial training further enables the generation of realistic high-resolution images, especially with complex structures. SRFS~\cite{0.8}, combining fast and slow scanning modes allows for more efficient detail recovery in SEM images. The de-noising diffusion model excels in detail recovery and noise reduction, providing practical value for specific microscopic image processing. SRCNN~\cite{dong2015image}, as introduced by Dong et al., is pioneering in incorporating a simple three-layer convolutional neural network for super-resolution, surpassing the performance of conventional techniques. Subsequently, Kim et al.~\cite{kim2016accurate} advance the field by proposing VDSR, a more deeply-stacked architecture that further enhances performance. This paves the way for increasingly intricate models like RDN~\cite{zhang2018residual}, substantiating the notion that wider and deeper networks yield superior reconstruction outcomes.
However, these complex models bring feature redundancy and high computational costs, making them less practical for real-world, resource-constrained applications. 
As such, there is growing demand for lightweight super-resolution networks that balance complexity and performance. 
%
Existing methods often rely on structural priors, which primarily focus on low-level details.
However, these methods lack explicit semantic priors, which are essential for fine detail recovery, and fail to fully utilize both local and non-local information, limiting their ability to restore fine-grained details and maintain global consistency.

\begin{figure*}[!t] 
\centering
\includegraphics[width=0.95\linewidth]{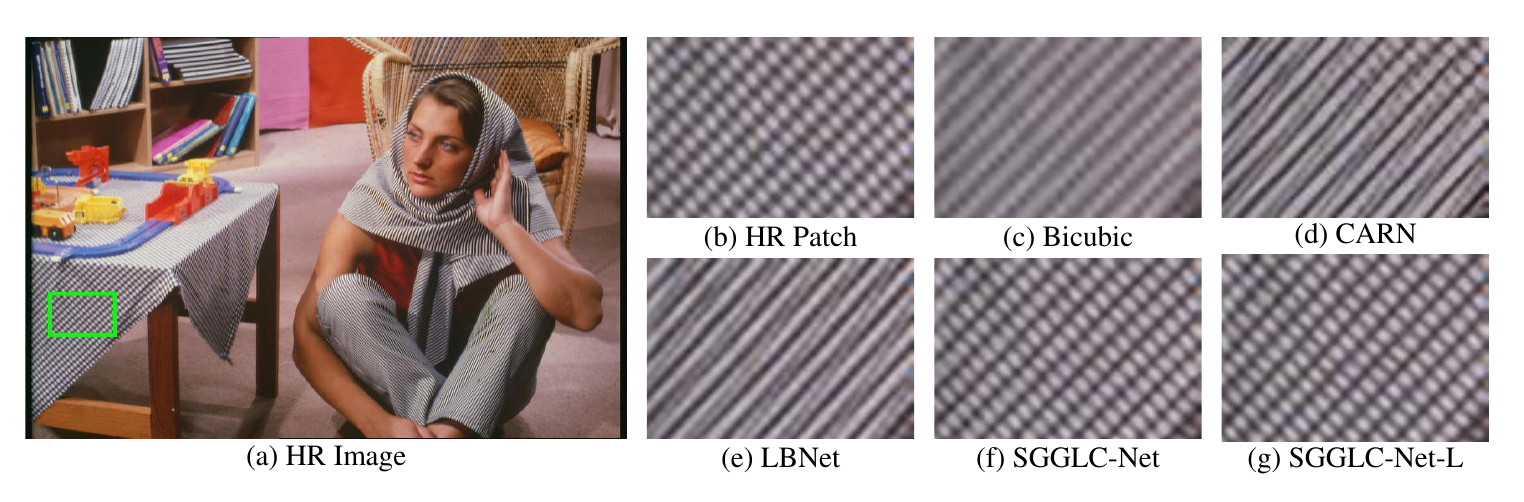}
\caption{Challenging super-resolution example. 
(a) and (b) are the HR image and HR patch. (c) is the result restored by bicubic interpolation. 
(d) and (e) are the restoration results of CARN~\cite{ahn2018fast} and LBNet~\cite{DBLP:conf/ijcai/GaoW0L0Z22}, which do not use semantic priors guidance. 
(f) and (g) are the restoration results of our SGGLC-Net and SGGLC-Net-L, which effectively exploit semantic priors guidance.
Our SGGLC-Net family is able to recover better details compared with previous state-of-the-art methods.
}
\label{fig:p1} 
\vspace{-4mm}
\end{figure*}

To address the above challenges, we propose a Semantic-Guided Global-Local Collaborative Network (SGGLC-Net) for lightweight single-image super-resolution (SISR), 
which incorporates semantic priors extracted from a pre-trained VGG19 model~\cite{VGG} to guide the super-resolution process. 
These semantic priors are directly integrated into the Semantic Guidance Module (SGM), which refines the features by employing attention mechanisms. 
By embedding semantic priors, SGM improves the network's ability to restore fine details while maintaining semantic consistency.
To further enhance the interactions between local and non-local information for improved detail rendition, we propose a Global-Local Collaborative Module (GLCM). 
The GLCM features three Global and Local Detail Enhancement Modules, followed by a Hybrid Attention Block, where both of them work together to efficiently learn more useful features for enhanced image
restoration. 
By integrating semantic priors through SGM and leveraging the global-local interactions of GLCM,
%
%
%
%
%
our SGGLC-Net offers a favorable balance among model parameters, computational demands, and performance metrics, as depicted in Fig.~\ref{fig:p7}.
Fig.~\ref{fig:p1} visually compares the quality of image restoration with and without the incorporation of semantic priors. It is evident that our semantically informed models retain more reliable edge and texture details, thereby achieving superior reconstruction quality over previous state-of-the-art approaches.

The main contributions of this paper as follows:
\begin{itemize}
\item We propose a Semantic-Guided Global-Local Collaborative Network (SGGLC-Net) for lightweight single-image super-resolution, which incorporates semantic priors into the SR task for better detail generation.
\item We design the Semantic Guidance Module (SGM) and the Global-Local Collaborative Module (GLCM), which collaboratively enhance super-resolution performance. 
SGM leverages semantic priors for feature refinement and fine detail recovery, while GLCM integrates global-local interactions to model features efficiently. Their collaboration ensures the effective use of semantic priors, enabling SGGLC-Net to restore complex textures and achieve superior performance with improved efficiency.
\item Extensive qualitative and quantitative experiments on
commonly used benchmarks show that our SGGLCNet
outperforms existing state-of-the-art lightweight SR
methods while achieving a better trade-off between model
complexity and super-resolution performance, demonstrating
its potential to enhance the effectiveness of visual
measurement systems.
\end{itemize}

\section{Related Work}
In this section, we review the SISR methods in Sec.~\ref{sec:Single-Image Super-resolution} and lightweight SISR approaches in Sec.~\ref{sec: Lightweight Single-Image Super-Resolution}.
We also introduce prior-guided methods in Sec.~\ref{sec: Prior-Guided Methods}.
%
\subsection{Single-Image Super-resolution}\label{sec:Single-Image Super-resolution}
In recent years, convolutional neural networks (CNNs) have excelled in single image super-resolution (SISR). 
Dong et al. ~\cite{dong2015image} pioneer CNN-based SISR with SRCNN, outperforming traditional methods. 
Further studies have showed that wider and deeper networks could improve reconstruction~\cite{kim2016accurate,zhang2018residual}, though often with computational redundancy. 
To address this, attention mechanisms have introduced. Zhang et al.~\cite{zhang2018image} present the RCAN, leveraging channel attention to enhance relevant features and improve quality. 
Subsequent models further optimized attention for efficiency and reconstruction, such as~\cite{DBLP:journals/tcsv/ZhuGRHHF22}, which employs expectation-maximization attention for visual quality with lower computational costs. 
In recent advances, 
Transformer-based networks have been widely applied in computer vision and achieved innovative reconstruction quality, which also promotes the advancement in the field of SISR. 
For example, Chen et al.~\cite{chen2021pre} propose a large pre-training model IPT to handle various low-level vision tasks. Liang et al.~\cite{liang2021swinir} introduce Swin Transformer~\cite{liu2021swin} to super-resolution tasks and propose SwinIR, which achieves excellent results. 
Although the aforementioned methods can achieve excellent reconstruction results, their extensive parameter and computation present significant challenges for their implementation in real-world scenarios.

%
\subsection{Lightweight Single-Image Super-Resolution}\label{sec: Lightweight Single-Image Super-Resolution}

A lightweight model refers to a network with fewer parameters, lower computational complexity, and less memory usage, making it more efficient for real-time applications.
To obtain a better balance between model parameters and performance, several methods are proposed to explore lightweight architectural designs.
Early lightweight super-resolution networks mainly reduce model parameters by recursive operations~\cite{kim2016deeply}. 
However, this approach only reduces the number of model parameters without effectively decreasing the computational complexity, failing to achieve real lightweight. 
%
%
Then, researchers further improve several multi-path network structures to make the network lightweight~\cite{yu2023dipnet,DBLP:journals/mlc/LiZLGW23}. 

Recent approaches combine CNNs with Transformers to leverage the strengths of both to improve the reconstruction performance. 
%
Gao et al.~\cite{DBLP:conf/ijcai/GaoW0L0Z22} introduce a hybrid network that uses CNNs for high-frequency extraction and Transformers for long-range information, enhancing detail preservation and performance.
%
Lu et al.~\cite{lu2022transformer} propose a hybrid CNN-Transformer network that uses a CNN module for high-frequency extraction and a Transformer for long-range information, enhancing detail preservation.
%
%

Despite the above methods having achieved good results, 
the limited depth of lightweight network structures still presents challenges in effectively recovering image details, particularly in image edge restoration. 
Further advancements are needed to improve their capability in restoring such complex features.

\subsection{Prior-Guided Methods}\label{sec: Prior-Guided Methods}
Due to the inherently ill-posed nature of SISR tasks, prior-based methods for single-image super-resolution can explicitly or implicitly utilize additional image priors to better recover high-frequency information lost in the image~\cite{liu2022dsrgan,wu2023learning,DBLP:journals/tip/QiLZGHS22}. 
For example, Liu et al.~\cite{liu2022dsrgan} design a detail prior to guide the network to restore the SR image, which is obtained by a model-based algorithm to provide more realistic details. 
%

Furthermore, several tasks have applied semantic priors to guide network learning, achieving good results and demonstrating the reliability of semantic priors. 
%
%
Wu et al.~\cite{wu2023learning} incorporate semantic knowledge into a low-light enhancement task and propose a novel semantic-aware knowledge-guided framework called SKF, which enhances color consistency and improves image restoration quality by learning rich priors encapsulated in semantic segmentation models. 
%
%
Qi et al.~\cite{DBLP:journals/tip/QiLZGHS22} introduce semantic prior features into the underwater image enhancement task to provide a corresponding realistic reference for the degraded images, enabling the model to better deal with unknown degradations.

In addition, there are also some works that apply semantic priors information to face super-resolution tasks. 
Zhao et al.~\cite{zhao2020saan} design a network called SAAN, which combines facial semantic priors with face super-resolution models and uses face prior information to guide the super-resolution network to obtain more realistic facial details through a semantic attention adaptation module. 
Wang et al.~\cite{wang2022propagating} propose a facial prior knowledge distillation FSR network called KDFSRNet. 
In specific terms, the teacher network extracts the facial structure information that the student network missed while avoiding the deficiency caused by prior estimation. 

However, currently, there are still relatively few works that utilize semantic priors information to guide lightweight super-resolution network learning. 
This work proposes a Semantic Guidance Module to extract more comprehensive semantic priors, guiding the super-resolution network to learn more useful image details for better image super-resolution.

\section{Proposed Approach}

%
%
%
%

Our goal is efficient super-resolution of low-resolution images. We propose SGGLC-Net, a lightweight network that leverages semantic priors from VGG19~\cite{VGG} to guide super-resolution. 
To integrate these priors, we introduce a Semantic Guidance Module (Sec.\ref{sec: Semantic Guidance Module}). 
For enhanced detail, a Global-Local Collaborative Module enables both local and non-local 
interactions (Sec.\ref{sec:Global-Local Collaborative Module}).

\begin{figure*}[t] 
\centering
\includegraphics[width=0.97\linewidth]{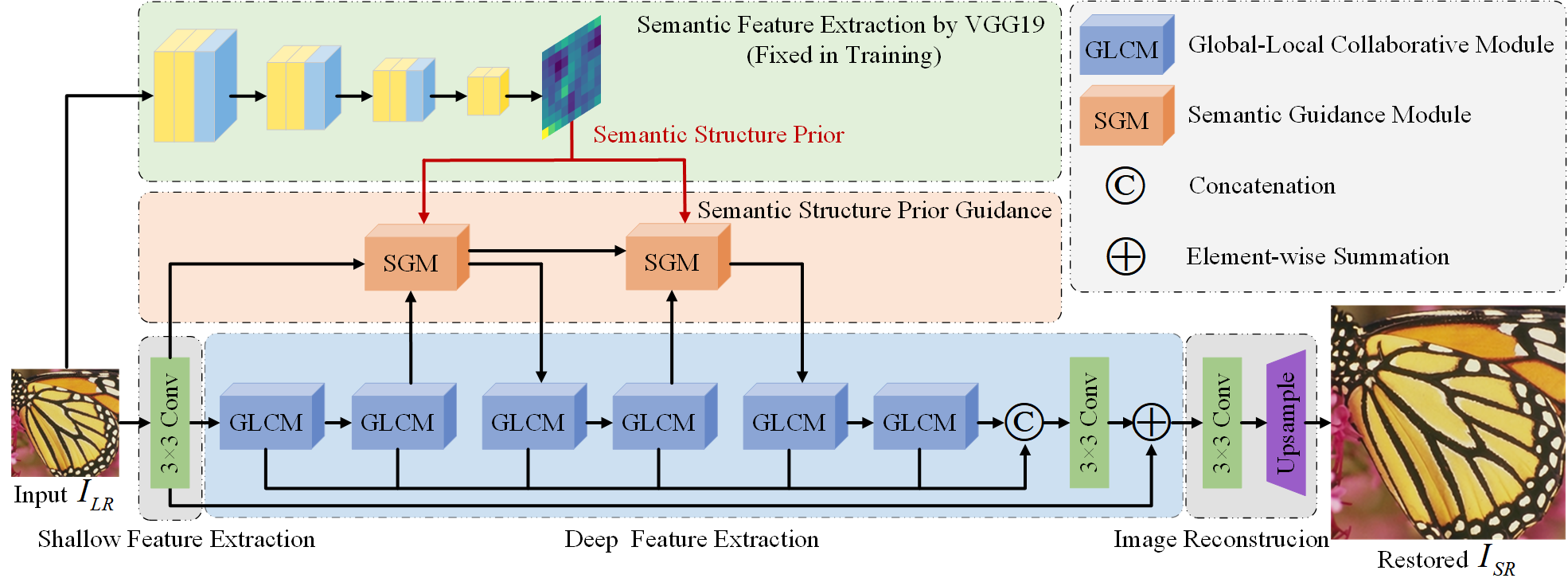}
\caption{The overall architecture of the proposed Semantic-Guided Global-Local Collaborative Network (SGGLC-Net). 
Our SGGLC-Net consists of four major components: Semantic Guided Module, Shallow Feature Extraction Module, Deep Feature Extraction Module, and Reconstruction Module.
Given a low-resolution (LR) input \(I_{LR}\), we first use a 3$\times$3 convolutional layer to extract shallow features. 
Then, we use a series of Global-Local Collaborative Modules (GLCM), as shown in Fig.~\ref{fig:p4}, as the Deep Feature Extraction Module to extract deeper features. 
Meanwhile, we also employ the Semantic Guidance Module (SGM), as shown in Fig.~\ref{fig:p3}, to effectively exploit the useful semantic priors obtained from the pre-trained extractor VGG19~\cite{VGG} to guide the process of deep feature extraction. 
Finally, we use a Reconstruction Module, which consists of a 3$\times$3 convolutional layer and a pixel shuffle layer to up-sample the fused feature to generate the super-resolution image \(I_{SR}\).
}
\label{fig:p2}
\end{figure*}

%
\subsection{Overall Pipeline}
Fig.~\ref{fig:p2} shows the overall pipeline of our Semantic-Guided Global-Local Collaborative Network (SGGLC-Net), which mainly consists of four components: Semantic Feature Guidance Module, Shallow Feature Extraction Module, Deep Feature Extraction Module, and Image Reconstruction Module. 

Given a low-resolution (LR) input \(I_{LR}\), we first use a 3$\times$3 convolutional layer to extract shallow features. 
Then, we use a series of Global-Local Collaborative Modules (GLCM), which is introduced in Sec.~\ref{sec:Global-Local Collaborative Module}, as the Deep Feature Extraction Module to extract deeper features. 
Meanwhile, we also employ the Semantic Guidance Module (SGM), which is detailed in Sec.~\ref{sec: Semantic Guidance Module}, to effectively exploit the useful semantic priors obtained from the pre-trained extractor VGG19~\cite{VGG} to guide the learning process of deep feature extraction. 
Similar to the previous works~\cite{zhang2018residual}, we use a global residual connection to fuse the shallow features and deep features.
Finally, we use a Reconstruction Module, which consists of a 3$\times$3 convolutional layer and a pixel shuffle layer to up-sample the fused feature to generate the super-resolution image \(I_{SR}\).

\subsection{Semantic Guidance Module}\label{sec: Semantic Guidance Module}
The pre-trained VGG19 network~\cite{VGG}, trained on the large-scale ImageNet dataset~\cite{deng2009imagenet} with object-level semantic labels, has been widely shown to capture hierarchical feature representations, ranging from low-level textures to high-level semantic features. 
Specifically, lower layers (e.g., Conv1-3) focus on edges and textures, while deeper layers (e.g., Conv4 and Conv5) encode object-level semantic information.
These high-level features are referred as semantic priors, provide rich contextual information beneficial for visual perception tasks, including image super-resolution. 
To harness this benefit, we extract semantic priors from VGG19~\cite{VGG} to guide the restoration process.
In particular, we select the \texttt{Conv4\_2} layer as the source of semantic priors, as it achieves a favorable balance between high-level semantic abstraction and fine-grained visual details, making it particularly suitable for super-resolution tasks.
%
\begin{figure}[t]
\centerline{\includegraphics[width=0.96\linewidth]{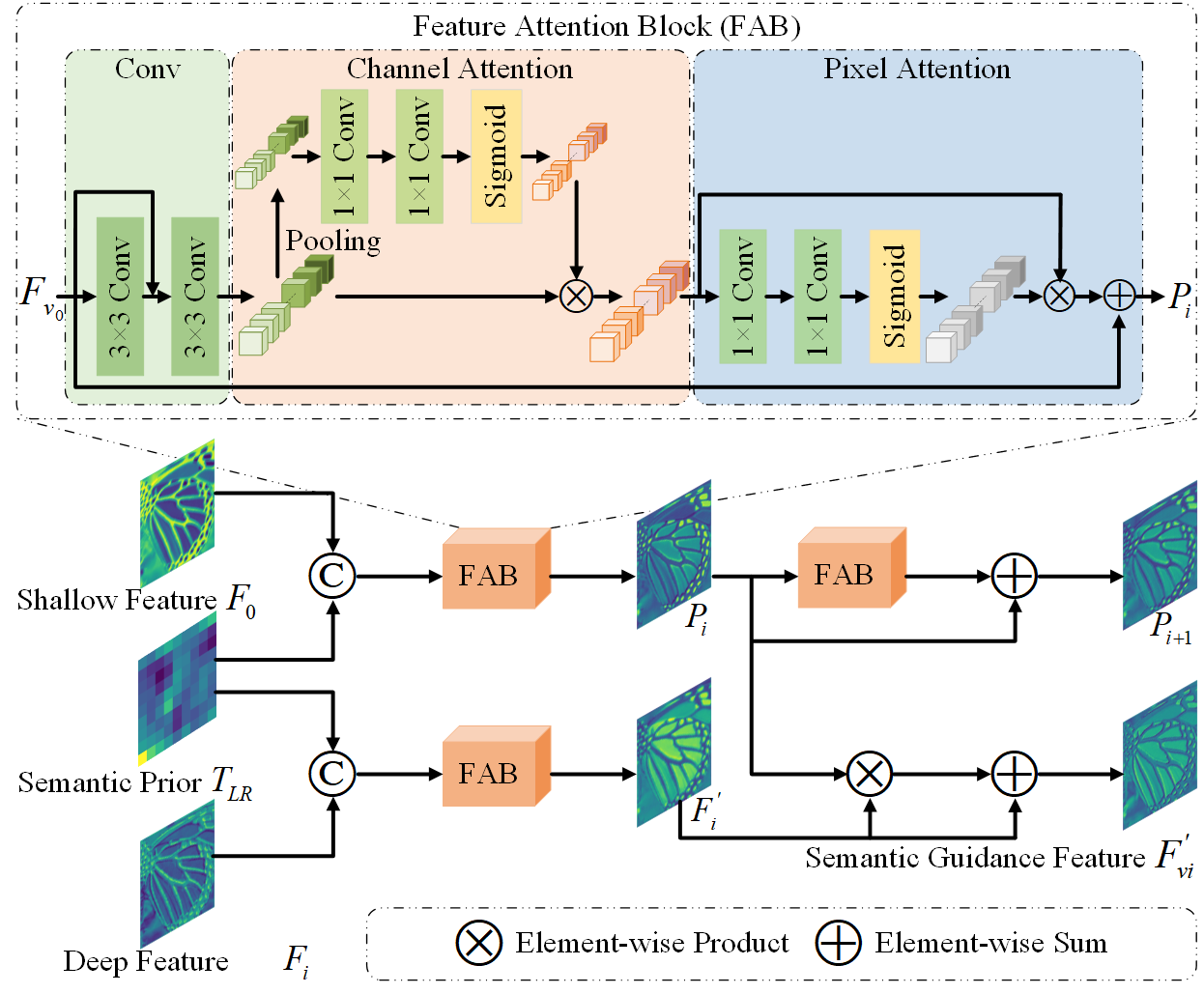}}	
\caption{The architecture of the Semantic Guidance Module (SGM). 
The SGM effectively utilizes the semantic priors extracted from a pre-trained VGG19 network~\cite{VGG} to guide the learning of deep feature extraction of the super-resolution backbone.
Note that \(F_{vi}^{'}\) is the input of the next GLCM while the \(P_{i+1}\) serves as one of the inputs of the next SGM.
}
\label{fig:p3}
\end{figure}

\begin{figure*}[!t] 
\centering
\includegraphics[width=0.97\linewidth]{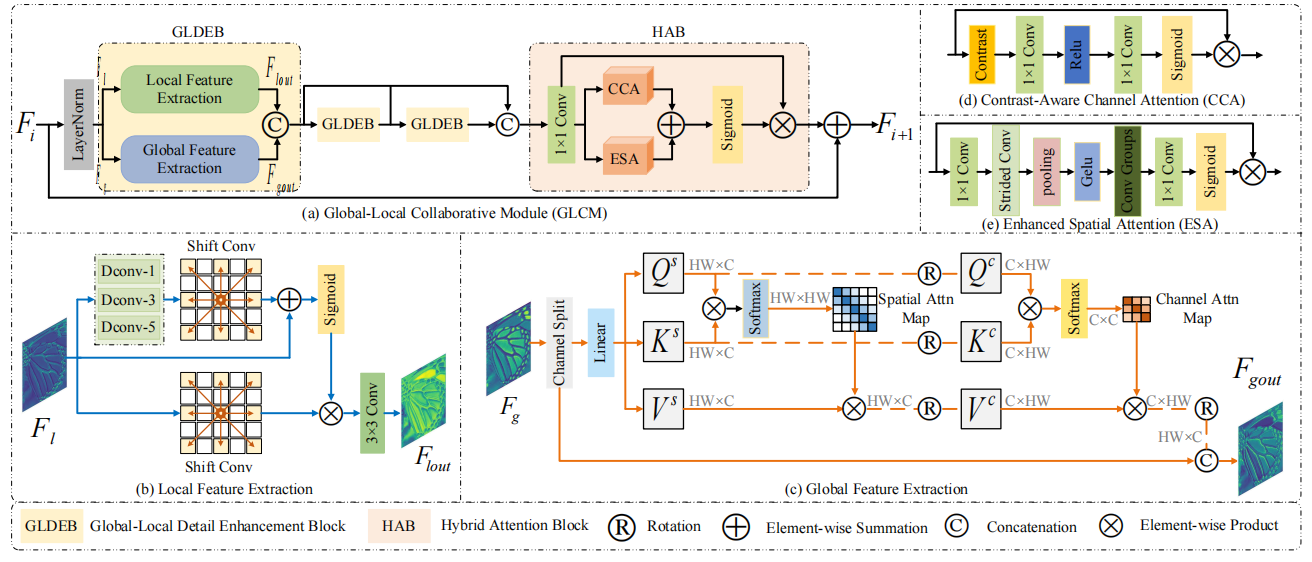}
\caption{{(a) The architecture of the proposed Global-Local Collaborative Module (GLCM). 
Our GLCM consists of 3 Global-Local Detail Enhancement Blocks (GLDEB) to interact with the local and non-local information to better super-resolve images, and a Hybrid Attention Block (HAB) to dynamically select more useful features for better image super-resolution.
(b) The architecture of the local feature extraction. 
(c) The structure of the global feature extraction.}}
\label{fig:p4}
\end{figure*}

Nevertheless, these semantic priors often include considerable redundant information. 
To mitigate this, we introduce a Semantic-Guided Module (SGM), which is comprised of several Feature Attention Blocks (FAB) inspired by SAAN~\cite{zhao2020saan} to help the network better capture and preserve fine image details and contextual semantic priors information.
The architecture of the SGM is illustrated in Fig.~\ref{fig:p3}.

Notably, we only utilize the semantic priors twice, which are the inputs to the third and fifth GLCMs to facilitate better image super-resolution since we notice that using semantic priors twice in the super-resolution network would produce better performance (see Sec.~\ref{sec: Effect on the Semantic Prior} for ablation analysis).  
To better exploit shallow features to adaptively learn more useful features, we also incorporate shallow features into the SGM.

We define \(F_0\) as the extracted shallow features, and \(F_i\) represents the output of the i-th GLCM. This feature flow process of SGM can be expressed as follows:
\begin{subequations}
\label{eq:GLCM}
\begin{align}
&\ F_{v_0} = Conv_{1\times 1} \left(  Concat \left[ F_0, T_{LR} \right] \right),
\\
&\ F_{v_i} = Conv_{1\times 1} \left(  Concat \left[ F_i, T_{LR} \right] \right),
\\
&\ P_i = f_{FAB} \left( F_{v_0} \right),
\\
&\ F_{i}^{'} = f_{FAB} \left( F_{v_i} \right),
\\
&\ F_{v_i}^{'} = F_{i}^{'} + F_{i}^{'} \times P_i,
\\
&\ P_{i+1} = P_i + f_{FAB} \left( P_i \right),
\end{align}
\end{subequations}
where \(T_{LR}\) represents the input semantic prior features; \(F_{v_0}\) represents the shallow features which would be fused with semantic priors; \(F_{v_i}\) represents the deep features fused with semantic priors; \(Conv_{1\times 1} \left( \cdot \right)\) is the 1$\times$1 convolution function; \(Concat\left(\cdot \right)\) is the concatenation operation; \(f_{FAB} \left( \cdot \right) \) represents the Feature Attention Block; \(P_i\) and \(F_{i}^{'}\) represents \(F_{v_0}\) and \(F_{v_i}\) extracted through FAB respectively;  \(F_{v_i}^{'}\) represents the output features of the SGM.

%
\subsection{Global-Local Collaborative Module}\label{sec:Global-Local Collaborative Module}

To fully utilize the extracted features from the SGM and better super-resolve images, we propose a Global-Local Collaborative Module (GLCM), as depicted in Fig.~\ref{fig:p4}(a).
GLCM adopts a task-specific dual-branch design, where a convolutional branch preserves spatially localized details, while a self-attention branch selectively enhances long-range dependencies and global semantic consistency.
Unlike the global self-attention mechanisms used in ViT, which focus purely on global feature interactions, our self-attention branch is specifically designed to work alongside convolutions, ensuring that both global context and fine-grained local details are effectively captured.
This global-local collaboration ensures both fine-grained texture preservation and global context modeling, making it particularly effective for lightweight super-resolution.
Our GLCM mainly incorporates $3$ Global-Local Detail Enhancement Blocks (GLDEB) and a Hybrid Attention Block (HAB).
GLDEB aims to enrich both the local and non-local interactions for better image super-resolution, while the HAB aims to reduce the feature redundancy to facilitate better restoration.
The learning process of GLCM can be represented as follows:
\begin{subequations}
\label{eq:GLCM}
\begin{align}
&\ F_{GL_0} = LN \left( F_i \right),
\\
&\ F_{GL_k} = f_{GLDEB}^{k} \left( F_{GL_{k-1}} \right), k=1,2,3,
\\
&\ F_{fusion} = Conv_{ 1 \times 1 } \left( Concat \left[ F_{GL_1}, F_{GL_2}, F_{GL_3} \right] \right),
\\
&\ F_{fusion}^{'} = f_{CCA} \left( F_{fusion}\right) + f_{ESA}\left( F_{fusion} \right),
\\
&\ F_{i+1} = F_{i} + \sigma \left( F_{fusion}^{'} \right) \times F_{fusion},
\end{align}
\end{subequations}
where \(F_{GL_0}\) represents the input of the first GLDEB; \(F_i\) and \(F_{i+1}\) represent the input and output of the i-th GLCM, respectively; \(F_{GL_k}\) represents the output of the k-th GLDEB; \(f_{GLDEB}^{k} \left( \cdot \right) \) represents the k-th GLDEB; \(LN \left(\cdot \right)\) represents the layer normalization~\cite{ba2016layer} operation used for training stability; \(F_{fusion}\) represents the fused features of the GLDEB's output; \(F_{fusion}^{'}\) represents the fused features after passing through the attention module; \(f_{CCA} \left( \cdot \right)\) and \(f_{ESA} \left( \cdot \right)\) represent Contrast-aware
Channel Attention and Enhanced Spatial Attention, respectively, which is inspired by ~\cite{li2022blueprint}; \(\sigma \left( \cdot \right)\) represents the Sigmoid operation. 

%
\subsubsection{Global-Local Detail Enhancement Block}
To extract more details for better lightweight image super-resolution, we propose a Global-Local Detail Enhancement Block (GLDEB) to effectively learn local features and global features by the local feature extraction branch and global feature extraction branch, respectively. Notable, to make the network more lightweight, we input the features from half of the channels into each of these two branches.

\noindent \textbf{Local Feature Extraction.} 
Fig.~\ref{fig:p4}(b) shows the local feature extraction, which is inspired by FENet~\cite{behjati2022frequency}. 
Firstly, we apply an average pooling layer to get a small area of \(F_l\) called \(F_{down}\).
Then, we use a bicubic interpolation to upsample the feature to the same size as \(F_l\) to obtain \(F_{up}\).

Next, by taking the difference between the up-sampled feature \(F_{up}\) and the input \(F_{l}\), we can obtain the high-frequency local features. 

Additionally, a convolutional kernel of a certain size can only extract features at a specific scale. 
Therefore, we construct parallel branches with depthwise convolution with kernel sizes of 1, 3, and 5 to extract rich multi-scale features, providing more details for image restoration. 
Subsequently, we fuse these multi-scale features through addition to obtain the \(F_{l}^{'}\).
To further reduce the parameter and computational complexity of the model, we adopt 1$\times$1 convolution instead of the original 3$\times$3 convolution for feature extraction. 
However, 1$\times$1 convolution has a smaller receptive field and poorer feature extraction capability. Therefore, we introduce a Shift Conv. 
We preserve a portion of channel features without any changes and divide the remaining channel features into eight groups for shift operations, and the arrows in Fig.~\ref{fig:p4}(b) point towards the direction of the feature movements, simulating the convolution process of a 3$\times$3 kernel. 

The process of local feature extraction can be expressed as:
\begin{equation}
\ F_{l_{out}} = Conv_{3 \times 3}\left( SC\left( F_l\right) \times \sigma \left(SC (F_l^{'})+F_l\right)\right),
\end{equation}
where \(F_{l_{out}}\) represents the output of the local feature extraction branch; \(SC\left(\cdot \right)\) represents the Shift Conv module.

\noindent \textbf{Global Feature Extraction.} 
Fig.~\ref{fig:p4}(c) shows the global feature extraction. 
Firstly, to reduce unnecessary computations, we adopt a channel-wise splitting to split the input into two halves:
\begin{equation}
\ F_{g_1} , F_{g_2} = Split \left( Conv_{1 \times 1} \left( F_{g} \right) \right),
\end{equation}
where \(F_{g_1}\) and \(F_{g_2}\) are used as inputs to the two global branches;
$Split \left( \cdot \right)$ denotes the split operation.
As self-attention operations are effective in capturing global information, we apply the Omni Self-Attention (OSA)~\cite{wang2023omni} to extract global information. 
The OSA module utilizes rotation along the spatial axis to model self-attention in both spatial and channel dimensions, enabling the discovery of more potential information.

\begin{table*}[t]
\centering
\caption{
Quantitative results on widely-used benchmarks, including Set5~\cite{bevilacqua2012low}, Set14~\cite{zeyde2012single}, BSD100~\cite{martin2001database}, Urban100~\cite{huang2015single}, Manga109~\cite{matsui2017sketch} in terms of PSNR and SSIM. The \textbf{best} and  \underline{the second} results are highlighted in bold and underlined, respectively.
Higher values of PSNR and SSIM indicate better performance. }
%
\renewcommand{\arraystretch}{1.25}
\begin{tabular}{l|c|c|c|ccccc}
\hline
\multirow{2}{*}{Method}&\multirow{2}{*}{Scale}  &\multirow{2}{*}{Param (K)} & \multirow{2}{*}{FLOPs (G)}    & Set5~\cite{bevilacqua2012low}        & Set14~\cite{zeyde2012single}         & BSD100~\cite{martin2001database}       & Urban100~\cite{huang2015single}     & Manga109~\cite{matsui2017sketch}     
\\ \cline{5-9}
           &&&& PSNR/SSIM    & PSNR/SSIM     & PSNR/SSIM    & PSNR/SSIM    & PSNR/SSIM  \\\hline
VDSR~\cite{kim2016accurate}      &   & 612.6&150  & 37.53/0.9587 & 33.03/0.9124  & 31.90/0.8960 & 30.76/0.9140 & 37.22/0.9750 \\
CRAN~\cite{ahn2018fast}      &    & 1592&222.8 & 37.76/0.9590 & 33.52/0.9166  & 32.09/0.8978 & 31.92/0.9256 & 38.36/0.9765 \\
IMDN~\cite{hui2019lightweight}      &    &694 &158.8 & 38.00/0.9605 & 33.63/0.9177  & 32.19/0.8996 & 32.17/0.9283 & 38.88/0.9774 \\
AWSRN-M~\cite{wang2019lightweight}   &    & 1063 &224.2& 38.04/0.9605 & 33.66/0.9181  & 32.21/0.9000 & 32.23/0.9294 & 38.66/0.9772 \\
SMSR~\cite{wang2021exploring}      &    & 985 &131.6 & 38.00/0.9601 & 33.64/0.9179  & 32.17/0.8990 & 32.19/0.9284 & 38.76/0.9771 \\
MAFFSRN-L~\cite{muqeet2020multi} & $\times$2 & 790  &154.4& 38.07/0.9607 & 33.59/0.9177  & 32.23/0.9005 & 32.38/09308  & -/-          \\
FENet~\cite{behjati2022frequency}     &  & 600 &77.9 & 38.08/0.9608 & 33.70/0.9184  & 32.20/0.9001 & 32.18/0.9287 & 38.89/0.9775 \\
LBNet~\cite{DBLP:conf/ijcai/GaoW0L0Z22}     &    & 731 &90.3& 38.05/0.9607 & 33.65/0.9177  & 32.16/0.8994 & 32.30/0.9291 & 38.88/0.9775 \\
DIPNet~\cite{yu2023dipnet}    &    & 527 &14.5& 37.98/0.9605 & 33.66/0.9192  & 32.20/0.9002 & 32.31/0.9302 & 38.62/0.9770 \\
MPFFA~\cite{DBLP:journals/mlc/LiZLGW23}     &    & 523&66.2  & 38.09/0.9607 & 33.70/0.9190  & 32.21/0.9000 & 32.34/0.9296 & 38.98/0.9767 \\
MICU~\cite{chen2024micu} & & 1580  & - & 37.93/0.9601 & 33.63/0.9170 & 32.17/0.8987 & 32.09/0.9271 & -/- \\
\hdashline
\textbf{SGGLC-Net (Ours)}    &    & 490  &45& \underline{38.16}/\underline{0.9633} & \underline{33.79}/\underline{0.9207}  & \underline{32.27}/\underline{0.9008} & \underline{32.57}/\underline{0.9319} & \underline{39.10}/\underline{0.9782} \\
\textbf{SGGLC-Net-L (Ours)}    &    & 870  &  50&\textbf{38.13}/\textbf{0.9631} & \textbf{33.88}/\textbf{0.9212}
& \textbf{32.28}/\textbf{0.9010} & \textbf{32.61}/\textbf{0.9327} & \textbf{39.05}/\textbf{0.9780} 
\\ \hline
VDSR~\cite{kim2016accurate}      &    & 666 &612.6  & 33.66/0.9213 & 29.77/0.8314  & 28.82/0.7976 & 27.14/0.8279 & 32.01/0.9340 \\
DRCN~\cite{kim2016deeply}      &    & 1774 &17974.3 & 33.82/0.9226 & 29.76/0.8311  & 28.80/0.7963 & 27.15/0.8276 & 32.31/0.9328 \\
CARN~\cite{ahn2018fast}      &    & 1592 &118.8 & 34.29/0.9255 & 30.29/0.8407  & 29.06/0.8034 & 28.06/0.8493 & 33.43/0.9427 \\
IMDN~\cite{hui2019lightweight}      &    & 703  &71.5 & 34.36/0.9270 & 30.32/0.8417 & 29.09/0.8046 & 28.17/0.8519 & 33.61/0.9445 \\
SMSR~\cite{wang2021exploring}     &    & 993  &150.5 & 34.40/0.9270 & 30.33/0.8412  & 29.10/0.8050 & 28.25/0.8536 & 33.68/0.9445 \\
MAFFSRN-L~\cite{muqeet2020multi} & $\times$3 & 807  &68.5 & 34.45/0.9277 & 30.40/0.8432  & 29.13/0.8061 & 28.26/0.8552 & -/-          \\
FENet~\cite{behjati2022frequency}     &    & 600 &90.5  & 34.40/0.9273 & 30.36/0.8422  & 29.12/0.8060 & 28.17/0.8524 & 33.52/0.9444 \\
LBNet~\cite{DBLP:conf/ijcai/GaoW0L0Z22}     &    & 736 &90.5  & \underline{34.47}/0.9277 & 30.38/0.8417  & 29.13/0.8061 & 28.42/0.8559 & 33.82/0.9460 \\
MPFFA~\cite{DBLP:journals/mlc/LiZLGW23}    &    & 523 &66.2  & \textbf{34.51}/0.9283 & 30.37/0.8424 & 29.13/0.8058 & 28.35/0.8550 & 33.85/0.9461 \\
MICU~\cite{chen2024micu} & & 1580 &- & 34.38/0.9274 & 30.35/0.8419 & 29.10/0.8048 & 28.14/0.8518 & -/- \\
\hdashline
\textbf{SGGLC-Net (Ours)}     &    & 497 &48.5  & 34.32/\underline{0.9290} & \underline{30.49}/\underline{0.8459}  & \underline{29.25}/\underline{0.8135} & \underline{28.44}/\underline{0.8581} & \underline{33.95}/\underline{0.9469} \\
\textbf{SGGLC-Net-L (Ours)}    & & 856  &54.3 & 34.40/\textbf{0.9298} & \textbf{30.55}/\textbf{0.8468}  & \textbf{29.29}/\textbf{0.8147} & \textbf{28.61}/\textbf{0.8616} & \textbf{34.12}/\textbf{0.9479} 
\\ \hline
VDSR~\cite{kim2016accurate}      &    & 665  &612.6 & 31.35/0.8838 & 28.01/0.7674  & 27.29/0.7251 & 25.18/0.7524 & 28.83/0.8809 \\
DRCN~\cite{kim2016deeply}      &    & 1774 &1797.3 & 31.53/0.8854 & 28.02/0.7670  & 27.23/0.7233 & 25.14/0.7510 & 28.98/0.8816 \\
CARN~\cite{ahn2018fast}      &    & 1592  &90.9& 32.13/0.8937 & 28.60/0.7806  & 27.58/0.7349 & 26.07/0.7837 & 30.42/0.9070 \\
IMDN~\cite{hui2019lightweight}      &    & 715  &40.9 & 32.21/0.8948 & 28.58/0.7811  & 27.56/0.7353 & 26.04/0.7838 & 30.45/0.9075 \\
AWSRN-M~\cite{wang2019lightweight}   &    & 1254 &56.7 & 32.21/0.8954 & 28.65/0.7832  & 27.60/0.7368 & 26.15/0.7884 & 30.56/0.9093 \\
SMSR~\cite{wang2021exploring}      &    & 1006 &135.5 & 32.12/0.8932 & 28.55/0.7808  & 27.55/0.7351 & 26.11/0.7868 & 30.54/0.9085 \\
MAFFSRN-L~\cite{muqeet2020multi} & $\times$4 & 830  &38.6 & 32.20/0.8953 & 28.62/0.7822  & 27.59/0.7370 & 26.16/0.7887 & -/-          \\
FENet~\cite{behjati2022frequency}     &    & 600  &74.3 & 32.24/0.8961 & 28.61/0.7818  & 27.63/0.7371 & 26.20/0.7890 & 30.46/0.9083 \\
LBNet~\cite{DBLP:conf/ijcai/GaoW0L0Z22}    &    & 742  &90.8 & \underline{32.29}/0.8960 & 28.68/0.7832  & 27.62/0.7382 & 26.27/0.7906 & \underline{30.76}/0.9111 \\
DIPNet~\cite{yu2023dipnet}    &    & 543 &14.9 & 32.20/0.8950 & 28.58/0.7811  & 27.59/0.7364 & 26.16/0.7879 & 30.53/0.9087 \\
MPFFA~\cite{DBLP:journals/mlc/LiZLGW23}    &    & 534 &43  & 32.36/0.8968 & \underline{28.69}/0.7835  & 27.61/0.7372 & 26.29/0.7902 & 30.71/0.9115 \\
DeWRNet~\cite{0.9} & & 28000 &- & \textbf{32.50}/0.8975 & 28.68/0.7823 & 27.60/0.7360 & -/- & -/- \\
PGAN~\cite{0.5} & &- & - & \textbf{31.03}/0.8798 & 27.75/0.8164 & 26.35/0.6926 & 25.47/0.9574 & -/- \\
MICU~\cite{chen2024micu} & & 1580  &-& 32.21/0.8945 & 28.65/0.7820 & 27.57/0.7359 & 26.15/0.7872 & -/- \\
\hdashline
\textbf{SGGLC-Net (Ours)}    &   & 506   &42& 32.24/\underline{0.8986} & 28.66/0.7838  & \underline{27.65}/\underline{0.7389} & \underline{26.33}/0.7929 & \underline{30.76}/\underline{0.9116} \\
\textbf{SGGLC-Net-L (Ours)}   &   & 868 &47  & 32.26/\textbf{0.8991} & \textbf{28.71}/\underline{0.7849}  & \textbf{27.69}/\textbf{0.7402} & \textbf{26.48}/\underline{0.7976} & \textbf{30.94}/\textbf{0.9140}\\ \hline
\end{tabular}
\label{tab:t1}
\vspace{-2.5mm}
\end{table*}
\begin{table}[!t]
\centering
\tablestyle{4pt}{1}
\scriptsize 
\renewcommand{\arraystretch}{1.5}
\caption{Inference times for 4$\times$ SR of different models, where low-resolution images are super-resolved to a size of 1280$\times$720. Lower inference times indicate higher computational efficiency.}
\begin{tabular}{c|c|c|c|c|c}
\hline
Method  & CRAN~\cite{ahn2018fast}  & IMDN~\cite{hui2019lightweight} & AWSRN-M~\cite{wang2019lightweight} & MPFFA~\cite{DBLP:journals/mlc/LiZLGW23} & Ours \\
\hline
Time (ms)  & 26.44  & 38.29 & 17.78 & 48.46 & \textbf{15} \\
\hline
\end{tabular}
\label{tab:t20}
\vspace{-5mm}
\end{table}
%
We remain one branch while applying global self-attention operation to the other branch. Afterwards, we merge the dual branch together:
\begin{subequations}
\label{eq:GLCM}
\begin{align}
&\ F_{g1}^{'} = f_{OSA} \left( F_{g1}\right),
\\
&\ F_{g_{out}} = Concat \left[ F_{g1}^{'} , F_{g2} \right],
\end{align}
\end{subequations}
where \(f_{OSA}\left( \cdot \right)\) represents the Omni Self-Attention module; \(F_{g_1}^{'}\) represents the extracted global features; \(F_{g_{out}}\) represents the output features of the global feature extraction branch.

\subsubsection{Hybrid Attention Block}
To reduce the feature redundancy of the network for better image super-resolution, we propose a Hybrid Attention Block (HAB) to extract the feature efficiently. Fig.~\ref{fig:p4}(a) shows our HAB, which mainly consists of Contrast-Aware Channel Attention (CCA) and Enhanced Spatial Attention (ESA) which is inspired by \cite{li2022blueprint}. 

\noindent \textbf{Contrast-Aware Channel Attention.}
The Contrast-aware Channel Attention (CCA) is improved of the channel attention \cite{hui2019lightweight} proposed for the lightweight SR task. The difference lies in CCA utilizing the mean and the summation of standard deviation as the contrast information to calculate the channel attention weights, rather than using the mean of each channel-wise feature.

\noindent \textbf{Enhanced Spatial Attention.}
The Enhanced Spatial Attention (ESA) first applies a 1$\times$1 convolution to reduce the channel dimension of the input features. Subsequently, a stride convolution and a stride max-pooling layer are used to down-sample the spatial size. After that, only a 3$\times$3 convolution is used instead of a group of convolutions to extract the feature with a lower computational cost. Then, an interpolation-based up-sampling method is applied to restore the original spatial size. Later, combined with a residual connection, a 1$\times$1 convolution is used to restore the channel size. 
Finally, an attention matrix is generated by applying a Sigmoid function and then element-wise multiplied with the original input feature.
Finally, we connect the CCA and ESA by element-wise sum by following a sigmoid function to dynamically select more useful features by element-wise multiplying with the input for better image super-resolution.

\begin{figure*}[!t] 
\centering
\includegraphics[width=0.972\linewidth]{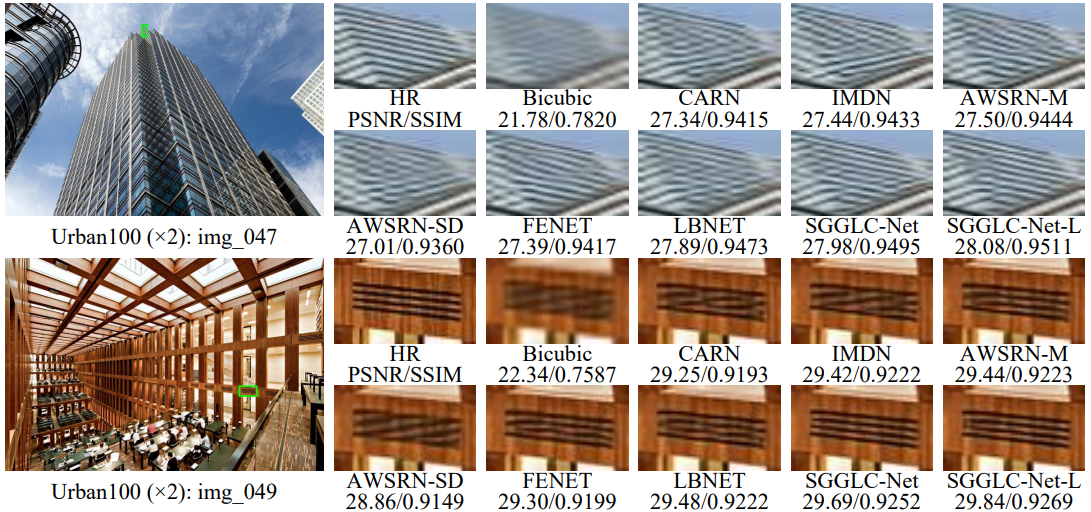}
\caption{Visual comparison with state-of-the-art lightweight SR methods on Urban100~\cite{huang2015single}, and areas of contrast are marked with green boxes on the original image. Among them, our SGGLC-Net family is able to produce much clearer results with better structural details.}
\label{fig:p5}
\vspace{-3mm}
\end{figure*}

\begin{figure*}[!t] 
\centering
\includegraphics[width=0.972\linewidth]{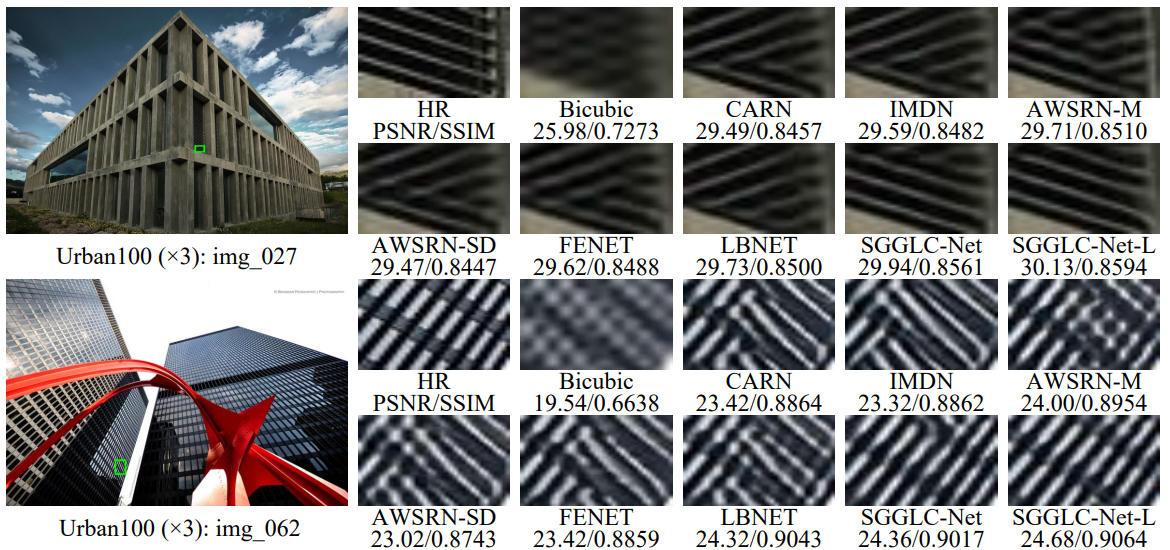}
\vspace{-1mm}
\caption{Visual comparison with state-of-the-art lightweight SR methods on Urban100~\cite{huang2015single}, and areas of contrast are marked with green boxes on the original image. Among them, our SGGLC-Net family is able to produce much clearer results with better structural details.}
\label{fig:p6}
\vspace{-2mm}
\end{figure*}
%
\subsection{Loss Function}
During the training process, for a given paired low-resolution/high-resolution data \(\left\{ I_{LR}^{i}, I_{HR}^{i}\right\} _{i=1}^{N}\), where \(N\) represents the number of image pairs, we use a \(L_1\) loss function to optimize the model:

\begin{equation}
\hat{\theta}= \mathop{\arg\min}\limits_{\theta} \frac{1}{N} \sum_{i=1}^{N} \Vert I_{SR}^{i} - I_{HR}^{i} \Vert_1,
\end{equation}
where \(\theta\) represents the trainable parameters; \(\Vert \cdot \Vert_1\) represents the \(L_1\) norm.

%

%
\section{Experiments}
In this section, we conduct extensive experiments to verify the effectiveness of the proposed SGGLC-Net.
\subsection{Dataset and Evaluation Metrics}

\noindent \textbf{Dataset.} Following previous works~\cite{hui2019lightweight,wang2019lightweight}, we train our SGGLC-Net using 800 images from the DIV2K~\cite{agustsson2017ntire} dataset and evaluate it on five publicly available datasets, including Set5~\cite{bevilacqua2012low}, Set14~\cite{zeyde2012single}, BSD100~\cite{martin2001database}, Urban100~\cite{huang2015single}, and Manga109~\cite{matsui2017sketch}. 
We also use the Real-World image super-resolution benchmark (RealSR)~\cite{DBLP:conf/iccv/CaiZYC019} to demonstrate the generalization capability of the proposed SGGLC-Net. 

\noindent \textbf{Evaluation Metrics.} We assess the effectiveness of our SGGLC-Net by using the Peak-Signal-to-Noise Ratio (PSNR) and Structural Similarity Index (SSIM)~\cite{wang2004image} as metrics to evaluate the image super-resolution quality, where higher values of PSNR and SSIM~\cite{wang2004image} indicate better performance. 
All the results are evaluated on the Y channel of the YCbCr color space.

\subsection{Implementation Details}
Following previous works~\cite{hui2019lightweight,wang2019lightweight}, we obtain LR images by using bicubic interpolation to downsample the HR image and enhance the training data by random rotation and horizontal flips.
During training, we randomly crop 64$\times$64 patches from the training set as the inputs.
We implement our SGGLC-Net on the PyTorch framework, optimized using the Adam optimizer~\cite{adam}, and trained on a single V100 GPU. 
We start with an initial learning rate of \(1 \times 10^{-3}\), which is halved every 200 epochs. 
The test time of our model only takes 15ms on the v100 GPU.
The number of channels of the model is set to 48, and there are 6 GLCMs. 
Additionally, we also introduce a larger version of the model, SGGLC-Net-L, with 64 channels. 
It is noteworthy that we only utilize the features extracted by VGG19~\cite{VGG} as prior inputs to the network, and the parameters of the VGG19~\cite{VGG} are not involved in the optimization process during training.
%
\subsection{Comparison with Lightweight SISR Models on Synthetic Benchmarks}
\subsubsection{Quantitative Results}
We compare our SGGLC-Net with 12 advanced lightweight SISR models, including VDSR~\cite{kim2016accurate}, DRCN~\cite{kim2016deeply}, CARN~\cite{ahn2018fast}, IMDN~\cite{hui2019lightweight}, AWSRN-M~\cite{wang2019lightweight}, SMSR~\cite{wang2021exploring}, MAFFSRN-L\cite{muqeet2020multi}, FENet~\cite{behjati2022frequency}, LBNet~\cite{DBLP:conf/ijcai/GaoW0L0Z22}, DIPNet~\cite{yu2023dipnet}, MPFFA~\cite{DBLP:journals/mlc/LiZLGW23}, DeWRNet \cite{0.9}, PGAN \cite{0.5} and MICU~\cite{chen2024micu} for upscaling by $\times$2, $\times$3, and $\times$4. 
Tab.~\ref{tab:t1} summarises the comparison results, where we can clearly observe that our SGGLC-Net achieves the best results with fewer model parameters.
Concretely, our method surpasses MPFFA~\cite{DBLP:journals/mlc/LiZLGW23}, DIPNet~\cite{yu2023dipnet} by 0.14dB, 0.17dB on Urban100~\cite{huang2015single} and 0.12dB, 0.16dB on Set14~\cite{zeyde2012single} for $\times$2 SR, and exceeds DIPNet~\cite{yu2023dipnet} by 0.17dB and 0.23dB on Urban100~\cite{huang2015single} and Manga109~\cite{matsui2017sketch} for $\times$4 SR, and also increase MPFFA~\cite{DBLP:journals/mlc/LiZLGW23} by 0.12dB, 0.12dB and 0.1dB on Set14~\cite{zeyde2012single}, BSD100~\cite{martin2001database} and Manga109~\cite{matsui2017sketch} for $\times$3 SR. 
Although the number of parameters of SMSR~\cite{wang2021exploring} is about 2 times than ours, the PSNR results of $\times$2, $\times$3, and $\times$4 SR are 0.24dB, 0.27dB and 0.22dB lower than ours on the Urban100~\cite{huang2015single}.
Moreover, our larger model SGGLC-Net-L can even bring further improvement.
We also provide the 4$\times$ SR inference times of several lightweight models in Tab.\ref{tab:t20}, which are selected from Tab.\ref{tab:t1} based on the availability of runtime data reported in their original papers and their relevance to lightweight super-resolution, demonstrating that our method achieves superior computational efficiency with faster inference times.
%
%
\subsubsection{Visual Comparison}
Meanwhile, we also provide the visual comparison between SGGLC-Net and other lightweight SISR models on $\times$2, $\times$3, and $\times$4 in Figs.~\ref{fig:p5} and \ref{fig:p6}. 
%
%
%
%
Obviously, our reconstruction results accurately preserve line directions, whereas other models produce incorrect or blurry lines, especially evident in the Urban100 dataset~\cite{huang2015single} with its rich texture information. This highlights the superiority of SGGLC-Net in restoring image structure and details.
Taking the image ``img049” in the Urban100 dataset~\cite{huang2015single} in Fig.~\ref{fig:p5} as an example, the stripes recovered by Bicubic, CARN~\cite{ahn2018fast}, IMDN~\cite{hui2019lightweight}, etc. 
%
%
For the image ``img062” in the Urban100 dataset~\cite{huang2015single} in Fig.~\ref{fig:p6}, the stripes recovered by other methods are either excessively blurred, or the direction of the stripes is wrong, while we recover the stripes in the right direction, and also much clearer. 
It is apparent that our model achieves a superior visual effect with a more realistic image structure and details.
However, existing lightweight SISR approaches always cannot effectively recover sharper structures.
%
For example, although CARN~\cite{ahn2018fast} consumes more parameters than our SGGLC-Net, it still hands down blurry textures.

%
\subsection{Comparison with Lightweight SISR Models on Real-World Benchmarks}
\begin{table*}[!t]
\centering
\caption{Comparison with advanced SISR methods on RealSR benchmark~\cite{DBLP:conf/iccv/CaiZYC019}. The \textbf{best} and  \underline{the second} results are highlighted in bold and underlined, respectively.
Higher values of PSNR and SSIM indicate better performance.} 
%
\begin{tabular}{c|cccc|cc}
\hline
\multirow{2}{*}{Scale} & IMDN~\cite{hui2019lightweight}         & ESRT~\cite{lu2022transformer}     &LK-KPN~\cite{DBLP:conf/iccv/CaiZYC019}        & MMSR~\cite{zhang2023lightweight}     &\textbf{ SGGLC-Net (Ours)}  &\textbf{ SGGLC-Net-L (Ours) } \\
\cline{2-7}
            &PSNR/SSIM &   PSNR/SSIM &   PSNR/SSIM     & PSNR/SSIM    & PSNR/SSIM    &     PSNR/SSIM    \\ 
\hline
$\times$3 & 30.29/0.857  &  30.38/0.857  & 30.60/0.863 &30.93/0.868 & \textbf{31.04}/\underline{0.869} & \underline{31.01}/\textbf{0.870} \\ 
\hdashline
$\times$4 & 28.68/0.815  & 28.78/0.815 &  28.65/0.820 & \underline{29.38}/\underline{0.830} & 29.35/\underline{0.830} & \textbf{29.43}/\textbf{0.832} \\ 
\hline
\end{tabular}
\label{tab:t6}
\vspace{-2mm}
\end{table*}
\begin{figure*}[!t] 
\centering
\includegraphics[width=0.97\linewidth]{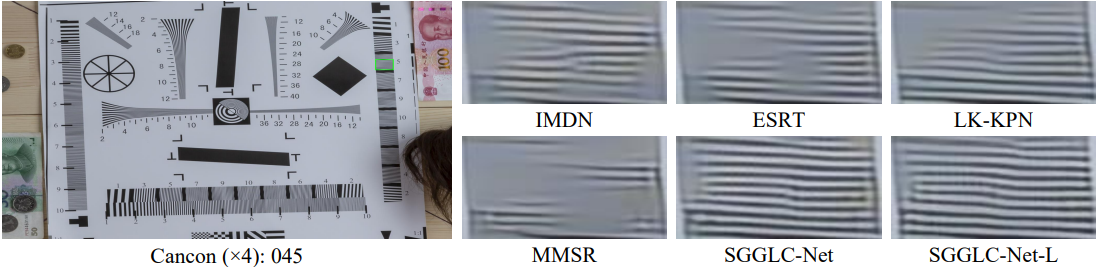}
\caption{Visual comparison on real-world images for $\times$4 on RealSR benchmark~\cite{DBLP:conf/iccv/CaiZYC019}.
Our SGGLC-Net family is able to generate much clearer super-resolution results with sharper details.
}
\label{fig:p12}
\vspace{-3mm}
\end{figure*}
To demonstrate the generalization capability of the proposed model, we conducted experiments on the Real-World image super-resolution dataset (RealSR)~\cite{DBLP:conf/iccv/CaiZYC019}. 
\subsubsection{Quantitative Results}
In Tab.~\ref{tab:t6}, we report the quantitative results compared with IMDN~\cite{hui2019lightweight}, ESRT~\cite{lu2022transformer}, LK-KPN~\cite{DBLP:conf/iccv/CaiZYC019} and MMSR~\cite{zhang2023lightweight}.
One can observe that our proposed SGGLC-Net-L achieves the best PSNR and SSIM~\cite{wang2004image} results, indicating its strong robustness to real-world scenarios. 
%
%
\subsubsection{Visual Comparison}
In Fig.~\ref{fig:p12}, we present qualitative comparison results, where we can observe that our proposed SGGLC-Net and SGGLC-Net-L can restore much clearer with sharper details. 
Notably, LK-KPN~\cite{DBLP:conf/iccv/CaiZYC019} is a dedicated network specifically designed for the real SR task.
However, our SGGLC-Net family still favors against LK-KPN~\cite{DBLP:conf/iccv/CaiZYC019}.
%

\begin{table}[!t]
\centering
\caption{Effect on the introduced semantic priors at different locations ($\times$2). 
The results demonstrate that embedding semantic priors twice yields the best performance, balancing the use of semantic priors with detail preservation.
The \textbf{best} results are highlighted in bold.}
\scalebox{0.87}{
\begin{tabular}{l|c|ccc}
 \hline
\multirow{2}{*}{Experiment}  &\multirow{2}{*}{Param (K)}   & Set14~\cite{zeyde2012single}  & Urban100~\cite{huang2015single}  & Manga109~\cite{matsui2017sketch}     \\
 \cline{3-5}

            &&     PSNR/SSIM     & PSNR/SSIM    & PSNR/SSIM        \\ 
\hline
(a) w/o VGG19 & 471 & 33.73/0.9201  & 32.28/0.9293 & 38.84/0.9776 \\
(b) w/ VGG19-1 & 476 & 33.70/0.9193 & 32.27/0.9295 & 38.87/0.9777 \\
\textbf{(c) w/ VGG19-2} & 490 &  \textbf{33.82}/\textbf{0.9206} & \textbf{32.48}/\textbf{0.9314} & \textbf{39.00}/\textbf{0.9779} \\
(d) w/ VGG19-6 & 499 & 33.67/0.9196  & 32.39/0.9305 & 38.92/0.9778 \\
(e) w/o FAB   & 490 & 33.75/0.9203  & 32.42/0.9309 & 38.93/0.9778 \\
\hline
\end{tabular}}
\label{tab:t2}
\end{table}
\begin{figure*}[!t] 
\centering
\includegraphics[width=0.97\linewidth]{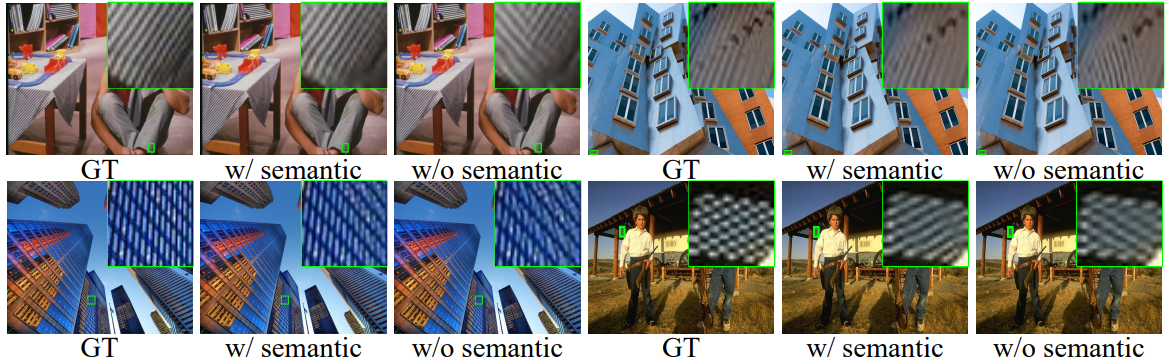}
\caption{The visual comparison between w/ and w/o semantic prior on Set14~\cite{zeyde2012single} and BSD100~\cite{martin2001database}, respectively.
Using semantic prior to guide a super-resolution network is able to help recover sharper details.
}
\label{fig:p11}
\end{figure*}
\begin{figure*}[!t] 
\centering
\includegraphics[width=0.97\linewidth]{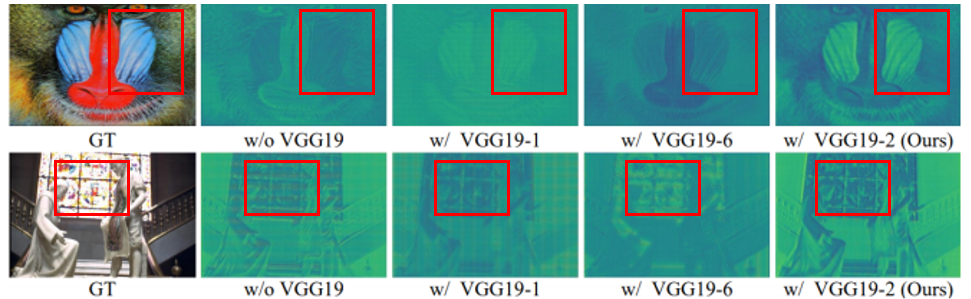}
\caption{The visualization of feature maps of average channels in SR images under different semantic prior settings on Set14~\cite{zeyde2012single} and BSD100~\cite{martin2001database}, respectively.}
\label{fig:p8}
\vspace{1mm}
\end{figure*}

\subsection{Ablation Study}
In this section, we conduct ablation experiments to investigate the effect of the proposed different components.
%
%
We test the PSNR and SSIM metrics~\cite{wang2004image} on the Set14~\cite{zeyde2012single}, Urban100~\cite{huang2015single}, and Manga109~\cite{matsui2017sketch}.
\subsubsection{Effect on the Semantic Priors}\label{sec: Effect on the Semantic Prior}
%
%

%
We first analyze the effect of semantic prior guidance within the following experiments.
%
\begin{itemize}
\item (a) w/o VGG19: No semantic priors is used in the model. 
\item (b) w/ VGG19-1: The semantic prior is utilized one time, which is input together with the input image. 
\item (c) w/ VGG19-2 (Our default): The semantic prior is utilized two times, which is input to the third and fifth GLCMs. 
\item (d) w/ VGG19-6: The semantic prior is utilized six times, which is input to all 6 GLCMs. 

\item (e) w/o FAB: Based on (c), the attention module is removed so that the semantic prior information is directly input to the GLCMs.
\end{itemize}
%
%
\begin{figure}[!h]\footnotesize
\begin{center}
\begin{tabular}{cccc}
\includegraphics[width=0.23\linewidth]{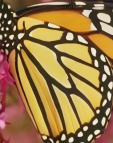} &\hspace{-4mm}
\includegraphics[width=0.23\linewidth]{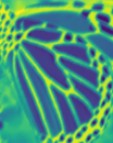} &\hspace{-4mm}
\includegraphics[width=0.23\linewidth]{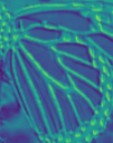} &\hspace{-4mm}
\includegraphics[width=0.23\linewidth]{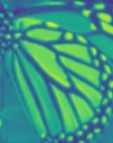} \\
(a)  & (b)    & (c) & (d)
\end{tabular}
\end{center}
\caption{Visual analysis of semantic prior integration in feature refinement.
(a) Input image. (b) Shallow features without semantic priors, capturing fine details but lacking global consistency. 
(c) Deep features without semantic priors, focusing on global context but losing details. 
(d) Refined features with semantic priors, effectively combining fine details with semantic coherence, enhancing feature representation.}
\label{fig:p1111}
\end{figure}


The experimental results summarized in Tab.~\ref{tab:t2} reveal notable performance improvements upon incorporating semantic priors (see Tab.~\ref{tab:t2}(a) vs. Tab.~\ref{tab:t2}(c)). 
When examining experiments detailed in Tab.~\ref{tab:t2}(b), (c), and (d), it becomes evident that optimizing semantic priors guidance twice during deep feature extraction yields the most effective results.
However, increasing the embeddings to six times (Tab.~\ref{tab:t2}(d)) results in diminishing returns, as excessive semantic prior information disrupts the balance between semantic priors and low-level detail recovery, leading to an overemphasis on abstractions and a loss of fine details, particularly in regions with complex patterns.
Additionally, a comparison between Tab.~\ref{tab:t2}(b) and (e) highlights the effectiveness of the FAB module~\cite{zhao2020saan}. 
By selectively refining the semantic priors, the FAB module improves their contribution to fine detail preservation and enhances overall image quality.
This is evident in the visual results shown in Fig.~\ref{fig:p11} and Fig.~\ref{fig:p8}, where two embeddings of semantic priors achieve better global consistency compared to six embeddings, which lose low-level details.  
These results confirm that a moderate amount of semantic priors optimally balances detail preservation and image quality.
To further investigate why semantic priors improve performance, 
we present a detailed visual analysis in Fig.~\ref{fig:p1111}  to illustrate how semantic priors refine feature representations.
Shallow features (Fig.~\ref{fig:p1111}(b)) emphasize local details such as edges but lacking global consistency. 
Deep features (Fig.~\ref{fig:p1111}(c)) capture coarse patterns and
global context but fail to preserve finer details. 
By integrating semantic priors, the refined features (Fig.~\ref{fig:p1111}(d)) combine fine details with global consistency, enabling a more balanced and consistent feature representation. 
This integration guides the network to recover both fine details and structural consistency during the super-resolution process.
\begin{table*}[t]
\centering
\caption{
Effect on the introduced SGM and GLCM ($\times$2). The \textbf{best} results are highlighted in bold.
Higher PSNR and SSIM values reflect improved performance.
The combination of GLCM and SGM achieves the highest PSNR/SSIM while maintaining a low parameter count, demonstrating the collaborative effects of the two modules.}
%
\begin{tabular}{l|ccccc}
\hline
\multirow{2}{*}{Experiment}     & Set5~\cite{bevilacqua2012low}        & Set14~\cite{zeyde2012single}         & BSD100~\cite{martin2001database}       & Urban100~\cite{huang2015single}     & Manga109~\cite{matsui2017sketch}      
\\ \cline{2-6}
            & PSNR/SSIM     & PSNR/SSIM      & PSNR/SSIM     & PSNR/SSIM     & PSNR/SSIM   \\\hline
(a) SGM+HAB~\cite{chen2023activating}            & 33.78/0.9619 & 33.26/0.9160  & 32.00/0.8977 & 31.41/0.9212 & 38.10/0.9759 \\
(b) GLCM+RCAB~\cite{zhang2018image}           & 38.11/0.9630& 33.72/0.9205  & 32.24/0.9005 & 32.48/0.9317 & 38.98/0.9779 \\
(c) GLCM+SGM           & \textbf{38.13/0.9631} & \textbf{33.88/0.9212}  & \textbf{32.28/0.9010} & \textbf{32.61/0.9327} & \textbf{39.05/0.9780} 
\\ \hline
\end{tabular}
\label{tab:t0.5}
\end{table*}

\begin{table*}[!t]
\centering
\caption{Ablation experiments of GLFEB ($\times$2). 
Note that we respectively input local feature extraction and global feature extraction branches with half of the channel's features.
While we disable the local feature extraction branch, i.e., (a), or global feature extraction branch, i.e., (e), in our method, the inputs are the features with all channels. 
Hence, (e) consumes more parameters since the local feature extraction branch contains more learning parameters.
Although this, our full model that uses both local feature extraction and global feature extraction branches achieves better super-resolution performance with a smaller model size than the model without a global feature extraction branch (e). The \textbf{best} results are highlighted in bold.
%
}
\scalebox{1}{
\begin{tabular}{l|c|ccc}
 \hline
\multirow{2}{*}{Experiment}       &\multirow{2}{*}{Param (K)}           & Set14~\cite{zeyde2012single}               & Urban100~\cite{huang2015single}     & Manga109~\cite{matsui2017sketch}     \\
 \cline{3-5}
            &&     PSNR/SSIM     & PSNR/SSIM    & PSNR/SSIM        \\ 
\hline
(a) w/o local feature extraction branch & 442 & 33.58/0.9184 & 32.13/0.9275 & 38.74/0.9774 \\
(b) w/o multipath in local feature extraction branch & 474  &   33.79/\textbf{0.9206} &  32.43/0.9310 & 38.93/0.9778 \\
(c) Shift Conv $\rightarrow$ Conv-1  in local feature extraction branch        & 467  & 33.71/0.9195 & 32.38/0.9303 & 38.93/0.9778 \\
(d) Shift Conv $\rightarrow$ Conv-3  in local feature extraction branch        & 633  & 33.75/0.9198  & \textbf{32.48}/0.9314 & 38.95/\textbf{0.9779} \\
\hdashline
(e) w/o global feature extraction branch      &  795 & 33.62/0.9186 & 32.41/0.9309 & 38.81/0.9775 \\
(f) w/o spatial attention in global feature extraction branch & 465  & 33.69/0.9197 & 32.36/0.9302 & 38.86/0.9776 \\
(g) w/o channel attention in global feature extraction branch &  459 & 33.73/0.9197 & 32.40/0.9306  & 38.85/0.9776 \\
\hdashline
(h) \textbf{Full (Ours)}          & 490  &  \textbf{33.82}/\textbf{0.9206} & \textbf{32.48}/\textbf{0.9315} & \textbf{39.00}/\textbf{0.9779}\\
\hline
\end{tabular}}
\label{tab:t3}
\end{table*}

%

\subsubsection{Effect on the GLCM and SGM}
%
To analyze the individual and combined impacts of the GLCM and SGM modules, we have conducted the following experiments:
\begin{itemize}
\item (a) SGM+HAB: 
Replacing GLCM with HAB~\cite{zhang2018image} to assess the standalone impact of SGM.
\item (b) GLCM+RCAB: 
     %
Replacing SGM with RCAB~\cite{chen2023activating} to isolate the standalone impact of GLCM.
\item (c) GLCM+SGM: 
Using both GLCM and SGM to evaluate their combined effects.
\end{itemize}

As illustrated in Tab.~\ref{tab:t0.5}, 
the combination of GLCM and SGM achieves the highest PSNR/SSIM across datasets while maintaining a low parameter count. 
These results highlight the complementary strengths of the two modules. 
SGM incorporates semantic priors to refine features and recover fine details, while GLCM leverages global-local interactions to effectively model features. 
Through its efficient design, GLCM minimizes the need for additional parameters, contributing to the lightweight nature of the framework. Together, these modules enable SGGLC-Net to achieve high-quality restoration with computational efficiency, confirming their collaborative design as a key contribution to superior super-resolution performance.

\subsubsection{Effect on the GLDEB}
Our GLDEB mainly consists of local feature extraction and global feature extraction. 
Among them, the local feature extraction is primarily composed of a multi-scale local feature extraction block and Shift Conv modeule, while the global feature extraction mainly consists of OSA~\cite{wang2023omni}. 
Therefore, we design the following experiments to demonstrate the effectiveness of each component in GLDEB.
\begin{itemize}
\item (a) w/o local feature extraction branch: Removing the local feature extraction, only retaining the global feature extraction to extract features from all channels.
\item (b) w/o multipath in local feature extraction branch: Removing the multi-scale module in local feature extraction, only retaining the branch with the scale of 1. 
\item (c) Shift Conv $\rightarrow$ Conv-1 in local feature extraction branch: Replacing Shift Conv with a 1$\times$1 convolution. 
\item (d) Shift Conv $\rightarrow$ Conv-3  in local feature extraction branch: Replacing Shift Conv with a 3$\times$3 convolution. 
\item (e) w/o global feature extraction branch: Removing the global feature extraction, only retaining the local feature extraction to extract all channel features.
\item (f) w/o spatial attention in global feature extraction branch: Removing the spatial attention in the global feature extraction. 
\item (g) w/o channel attention in global feature extraction branch: Removing the channel attention in the global feature extraction.
\item  (h) Full (Ours): our full model with the whole local feature extraction branch and global feature extraction branch.

\end{itemize}

As illustrated in Tab.~\ref{tab:t3}, several key findings emerge. Firstly, omitting local feature extraction results in a substantial drop in performance—by 0.24dB, 0.35dB, and 0.26dB on Set14~\cite{zeyde2012single}, Urban100~\cite{huang2015single}, and Manga109~\cite{matsui2017sketch} datasets, respectively. This suggests that capturing local features is essential for restoring image details, which is compromised with global feature extraction alone.
Secondly, the introduction of a multipath module yields a modest performance gain of 0.07dB on the Manga109~\cite{matsui2017sketch} dataset while incurring minimal additional parameters. This improvement is attributed to the module's capability to extract features from varying receptive fields, thereby enhancing image restoration.
Thirdly, employing shift convolution techniques significantly boosts performance by 0.11dB, 0.10dB, and 0.07dB on the Set14~\cite{zeyde2012single}, Urban100~\cite{huang2015single}, and Manga109~\cite{matsui2017sketch} datasets, respectively, demonstrating that shift operations can augment feature extraction without bloating the model's parameters.
%
Furthermore, we also validate the efficacy of our custom Shift Conv module by comparing it with a conventional 3$\times$3 convolution. Our module achieves superior results and maintains a smaller parameter size by emulating the receptive field of a 3$\times$3 convolution with a smaller kernel size through shift operations.
Fourthly, omitting global feature extraction and only applying local feature extraction performance decreased by 0.2dB and 0.19dB on Set14~\cite{zeyde2012single} and Manga109~\cite{matsui2017sketch} datasets, respectively, with a significant increase in the number of parameters. 
This indicates that global features are essential for image restoration, and relying solely on local feature extraction can compromise the process and impose a significant parameter burden.
%
Finally, applying spatial and channel attention in the global feature extraction branch improves performance by $0.14$ dB and $0.15$ dB on the Manga109~\cite{matsui2017sketch} dataset, respectively, demonstrating their importance for image super-resolution.

\subsubsection{Effect on the Number of GLCMs}
%

\begin{figure}[!t] 
\includegraphics[width=0.98\linewidth]{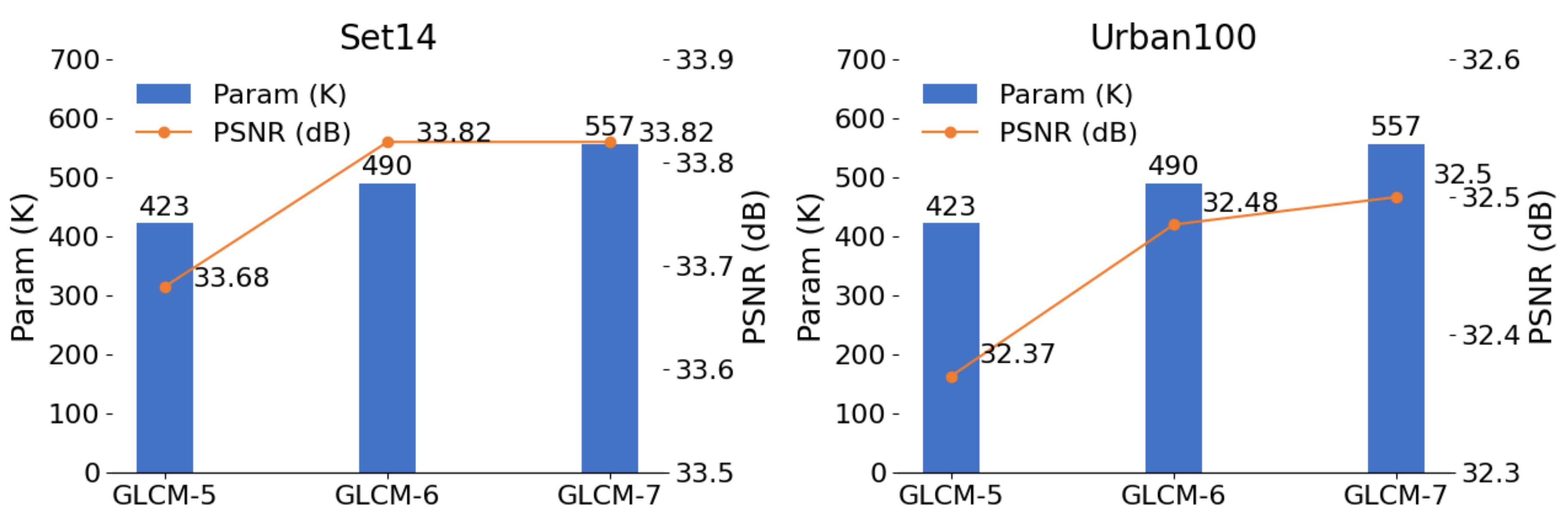}
\caption{The results ($\times$2) on different numbers of the Global-Local Collaborative Module (GLCM) on Set14~\cite{zeyde2012single} and Urban100~\cite{huang2015single}, respectively.}
\label{fig:p9}
\end{figure}

To optimize performance while considering model size, we experiment with varying the number of GLCMs in our network. We test configurations with 5, 6, and 7 GLCMs, as presented in Tab.~\ref{tab:t4} and Fig.~\ref{fig:p9}. Results indicate that while performance improves with increased network depth, the gains diminish beyond a certain point, leading to inefficient use of parameters. Thus, we opt for a configuration with 6 GLCMs to balance performance and model complexity.

\subsubsection{Effect on the Number of GLDEBs}
\begin{figure}[!t]
\includegraphics[width=0.96\linewidth]{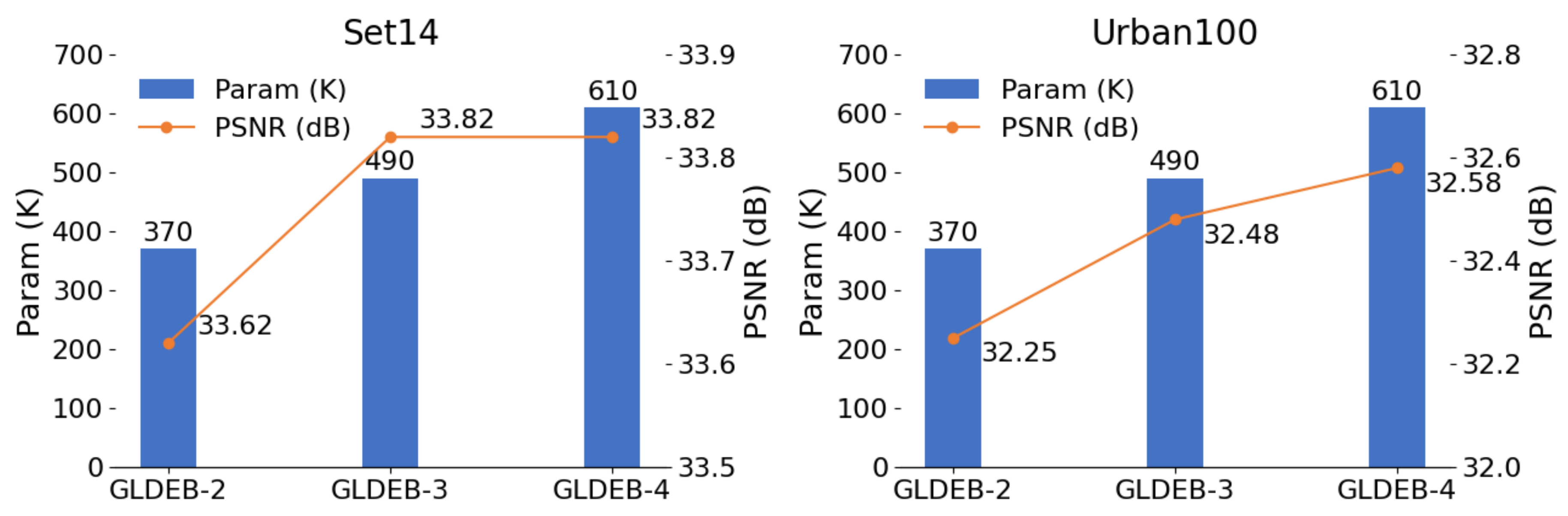}
\caption{The results ($\times$2) on different numbers of Global-Local Detail Enhancement Block (GLDEB) on Set14~\cite{zeyde2012single} and Urban100~\cite{huang2015single}, respectively. }
\label{fig:p10}
\vspace{5mm}
\end{figure}

%
%
To strike a balance between model size and performance, we vary the number of GLDEBs in our model and test configurations with 2, 3, and 4 GLDEBs. Results displayed in Tab.~\ref{tab:t5} and Fig.~\ref{fig:p10} show that deeper networks yield improved performance. Notably, in the Set14~\cite{zeyde2012single} and Urban100~\cite{huang2015single} datasets, increasing the number of GLDEBs led to substantial gains of 0.23dB and 0.09dB, respectively. Taking both performance and model complexity into account, we choose a configuration with 3 GLDEBs.

\begin{table}[!t]
\centering
\caption{Effect on the number of GLCM ($\times$2). The \textbf{best} results are highlighted in bold.}
\scalebox{0.905}{
\begin{tabular}{l|c|ccc}
 \hline
\multirow{2}{*}{Experiment}       &\multirow{2}{*}{Param (K)}          & Set14~\cite{zeyde2012single}                & Urban100~\cite{huang2015single}     & Manga109~\cite{matsui2017sketch}     \\
\cline{3-5}
            &&     PSNR/SSIM     & PSNR/SSIM    & PSNR/SSIM        \\ 
  \hline
(a) GLCM-5 & 423  & 33.68/0.9194 &  32.37/0.9303 & 38.92/0.9778 \\
\textbf{(b) GLCM-6} & 490   & \textbf{33.82}/0.9206 &  32.48/0.9315 & 39.00/0.9779 \\
(c) GLCM-7 & 557  &\textbf{33.82}/\textbf{0.9212} &  \textbf{32.50}/\textbf{0.9316} & \textbf{39.02}/\textbf{0.9780}\\
 \hline
\end{tabular}}
\label{tab:t4}
\end{table}

\subsubsection{Effect on the Different Deature Extractors}
To investigate the impact of different feature extractors on super-resolution, we evaluated ViT~\cite{vit}, CLIP~\cite{clip}, ResNet50~\cite{resnet50}, and VGG19 under the same conditions. 
The results summarized in Tab.~\ref{tab:t0.6} show that VGG19 achieves comparable or even slightly better PSNR/SSIM scores than more complex models, despite having significantly fewer parameters. 
Recent feature extractors such as ViT~\cite{vit} and CLIP~\cite{clip} are optimized for high-level semantic understanding in tasks like classification and vision-language tasks. 
These models focus on global semantic abstractions but often sacrifice fine-grained details critical for pixel-level reconstruction in super-resolution. Their increased complexity can lead to redundant or irrelevant features, negatively impacting detail recovery and structural consistency. 
In contrast, VGG19's hierarchical architecture effectively balances low-level detail extraction with high-level feature representation, making it particularly suitable for super-resolution tasks. VGG19's simplicity and efficiency provide an ideal trade-off between global consistency, detail preservation, and computational complexity.
\begin{table}[!t]
\centering
\caption{Effect on the number of GLFEB ($\times$2). The \textbf{best} results are highlighted in bold.}
\scalebox{0.89}{
\begin{tabular}{l|c|ccc}
 \hline
\multirow{2}{*}{Experiment}       &\multirow{2}{*}{Param (K)}          & Set14~\cite{zeyde2012single}             & Urban100~\cite{huang2015single}     & Manga109~\cite{matsui2017sketch}     \\
\cline{3-5}
            &&     PSNR/SSIM     & PSNR/SSIM    & PSNR/SSIM        \\ 
 \hline
(a) GLDEB-2 & 370  &  33.62/0.9194  & 32.25/0.9293 & 38.89/0.9777 \\
\textbf{(b) GLDEB-3} & 490  & \textbf{33.82}/\textbf{0.9206} &  32.48/0.9315 & 39.00/\textbf{0.9779} \\
(c) GLDEB-4 & 610  &  \textbf{33.82}/\textbf{0.9206} & \textbf{32.58}/\textbf{0.9324 }& \textbf{39.03}/\textbf{0.9779}\\
 \hline
\end{tabular}}
\label{tab:t5}
\end{table}

\begin{table}[!t]
    \centering
    \caption{Comparisons of different feature extractors in super-resolution performance.
    VGG19 achieves the best balance between performance and parameter efficiency, with higher PSNR and SSIM indicating better results.}
    \begin{tabular}{l|c|cc}
     \hline
    \multirow{2}{*}{Experiment}   &\multirow{1}{*}{Param}    & Set14~\cite{zeyde2012single}             & Urban100~\cite{huang2015single}          \\
    \cline{2-4}
                &  (K)&     PSNR/SSIM     & PSNR/SSIM           \\ 
     \hline
     (a) Vit Encoder &6327& 33.60/0.9184 &  32.12/0.9281    \\
    (b) CLIP Image Encoder&5982&  33.71/0.9189  & 32.33/0.9299  \\
    (c) Resnet50   &2693   & 33.53/0.9180 &  32.07/0.9273 \\
    (d) VGG19 &490&  \textbf{33.72}/\textbf{0.9198} & \textbf{32.37}/\textbf{0.9307} \\
     \hline
    \end{tabular}
    \label{tab:t0.6}
\end{table}

\subsubsection{Model Complexity Comparisons}
%

An analysis of the model's complexity is essential for a comprehensive evaluation. As shown in Fig.~\ref{fig:p7}, a comparative study of multi-add operations among various models is presented. Our SGGLC-Net strikes an optimal balance between model size, performance, and computational complexity as measured by multi-adds.
Specifically, our SGGLC-Net ranks second in performance on the Urban100~\cite{huang2015single} dataset while boasting the fewest model parameters. Moreover, our extended version, SGGLC-Net-L, delivers the top performance in the same setting. When compared with SMSR~\cite{wang2021exploring}, SGGLC-Net demonstrates superior reconstruction performance, despite having a similar number of parameters and multi-add operations.
%
In addition, although SGGLC-Net has slightly higher multi-add operations than LBNet~\cite{DBLP:conf/ijcai/GaoW0L0Z22}, it delivers better reconstruction results with fewer parameters, highlighting its efficiency and lightweight design in SISR.
%

\subsection{Limitations}
%

Although the pre-trained VGG19 network~\cite{VGG} has proven effective in extracting semantic prior information for enhancing image detail recovery, it introduces some limitations. Despite its lightweight nature compared to more complex models, the architecture can still lead to relatively higher memory usage and inference times, especially in large-scale applications, limiting its suitability for real-time or low-power scenarios.
%
%
%
To address these limitations, future work could explore replacing VGG19 with more efficient architectures, such as MobileNet, or utilizing techniques like model pruning and quantization to reduce computational overhead while maintaining performance.
%


%


\begin{figure}[!t] 
\centering
\includegraphics[width=0.95\linewidth]{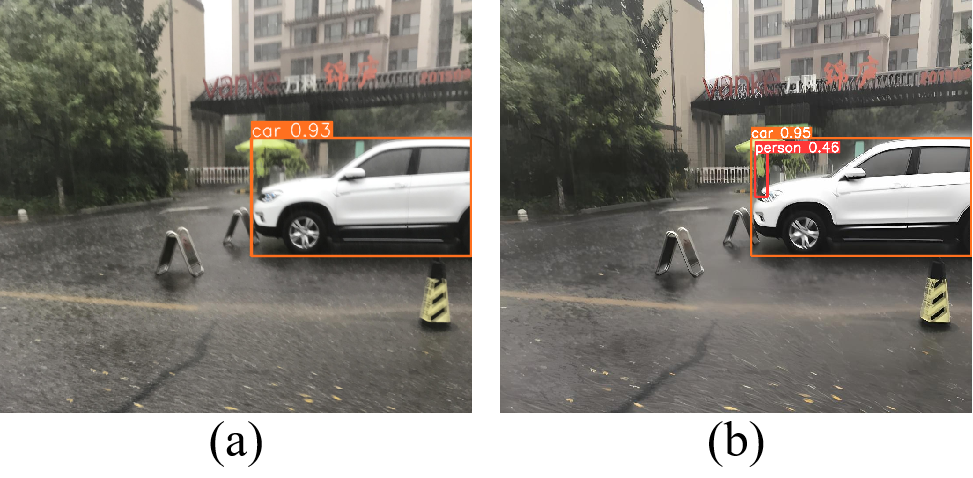}
    \caption{Visual comparisons of object detection before and after the super-resolution.
    (a) Object recognition result in the real-world image. (b) Object detection result after super-resolution enhancement using our proposed SGGLC-Net.
    }
\label{fig:p111}
\end{figure}

\subsection{Application}

%
%
%

To demonstrate the potential of SGGLC-Net in enhancing vision-based applications, we conduct experiments to evaluate its effectiveness in visual measurement tasks. Specifically, we use the yolov10 \cite{yolv} model to assess the super-resolution results obtained from our method. The evaluation is performed on the RID \cite{RID} dataset, which is used for object detection.
%
%
As illustrated in Fig.~\ref{fig:p111}, the super-resolved image achieves more precise and detailed object boundaries than the low-resolution image, particularly in recognizing individuals and cars. 
Notably, people undetectable in the low-resolution image are successfully identified, and the recognition accuracy of cars is significantly improved. These results demonstrate that SGGLC-Net effectively enhances visual data quality, boosting detection and recognition performance in measurement applications.

\section{Conclusion}
%
%

In this paper, we have proposed a Semantic-Guided Global-Local Collaborative Network (SGGLC-Net) designed specifically for enhancing vision-based measurement systems through lightweight single-image super-resolution. Our approach leverages semantic priors from a pre-trained model, such as VGG19~\cite{VGG}, to improve image quality and detail recovery, which is crucial for accurate measurement and analysis in various instrumentation and measurement applications.
%
%
%
%
%
%
Our SGGLC-Net employs a Semantic Guidance Module to integrate semantic prior information, and a Global-Local Collaborative Module to enhance both local and global feature interactions. This combination effectively addresses the challenges of image quality and detail preservation in measurement contexts.
%
Extensive experiments demonstrate that SGGLC-Net significantly outperforms existing methods in measurement contexts, improving both accuracy and reliability. 
Future work will extend our approach to other measurement and monitoring tasks, with further integration of semantic priors to support diverse instrumentation applications.

\bibliographystyle{IEEEtran}
\bibliography{sample-new}

\end{document}